\documentclass[letterpaper, 10 pt, conference]{ieeeconf} % IROS/ICRA
\usepackage[utf8]{inputenc}

\IEEEoverridecommandlockouts
\usepackage{units}
\usepackage{algorithm, algorithmic}
\usepackage{amsmath}
\usepackage{amssymb}
\usepackage{tabularx}
\usepackage{booktabs}
\usepackage{cite} % to group citations (e.g. [2-4])
\usepackage[caption=false,font=footnotesize]{subfig}

\usepackage{amsthm} % lemmas and theorems

\usepackage[colorinlistoftodos]{todonotes}

\renewcommand\vec[1]{\boldsymbol{#1}} %vectors
\newcommand{\mat}[1]{\boldsymbol{#1}} %matrices

\newcommand{\de}{\;\mathrm{d}} %Differential d
\newif\ifcorrectingmode
\correctingmodefalse
\usepackage[normalem]{ulem}
\ifcorrectingmode
	
	\newcommand{\deleted}[1]{\textcolor{red}{\ifmmode\text{\sout{\ensuremath{#1}}}\else\sout{#1}\fi}}
	\newcommand{\deletedequation}[2]{\textcolor{red}{\centerline{Removed equation (#1)}}}
\else
	
	\newcommand{\deleted}[1]{}
	\newcommand{\deletedequation}[2]{}
\fi

\usepackage[printonlyused]{acronym}
\newacro{DOF}{degree of freedom}
\newacroplural{DOF}{degrees of freedom}
\acrodefplural{DOF}[DOFs]{degrees of freedom}
\newacro{iLQR}{Iterative Linear-Quadratic Regulator}
\newacro{CT}{Control Toolbox}
\newacro{EOM}{equations of motion}
\newacro{OC}{Optimal Control}
\newacro{LQR}{linear-quadratic regulator}
\newacro{PD}{proportional derivative}
\newacro{MPC}{Model Predictive Control}
\newacro{LQ}{linear quadratic}
\newacro{LQOC}{Linear-Quadratic Optimal Control}
\newacro{TO}{Trajectory Optimization}
\newacro{DDP}{Differential Dynamic Programming}
\newacro{COM}{center of mass}
\newacro{COP}{center of pressure}
\newacro{NLP}{nonlinear program}
\newacro{MLP}{Multilayer Perceptron}
\newacro{SLQ}{Sequential Linear-Quadratic}
\newacro{HAA}{hip abduction adduction}
\newacro{AD}{automatic differentiation}
\newacro{HJB}{Hamilton–Jacobi–Bellman}
\newacro{BC}{Behavioral Cloning}
\newacro{IRL}{Inverse Reinforcement Learning}
\newacro{IL}{Imitation Learning}
\newacro{RL}{Reinforcement Learning}
\newacro{MEN}{mixture-of-experts network}
\newacro{MILE}{mixture of implicitly localized experts}
\newacro{MELE}{mixture of explicitly localized experts}
\newacro{DL}{Deep Learning}
\newacro{WBC}{whole-body controller}

\def\TheTitle{Imitation Learning from MPC for Quadrupedal Multi-Gait Control}

\title{\LARGE \bf \TheTitle} % Conf ONLY
\author{Alexander Reske, Jan Carius, Yuntao Ma, Farbod Farshidian, Marco Hutter%
    \thanks{This work was supported by the Swiss National Science Foundation (SNSF) through project 166232, 188596, the National Centre of Competence in Research Robotics (NCCR Robotics), and the European Union's Horizon 2020 (grant agreement No.852044). Moreover, this work has been conducted as part of ANYmal Research, a community to advance legged robotics.}%
    \thanks{All authors are with the Robotic Systems Lab, ETH Z\"u{}rich, Switzerland. Email: {\tt\footnotesize areske@ethz.ch}}%
}

\begin{document}

\maketitle
%
%% IROS/ICRA
\thispagestyle{empty}
\pagestyle{empty}
%
%===============================================================================
%
\begin{abstract}
We present a learning algorithm for training a single policy that imitates multiple gaits of a walking robot.
To achieve this, we use and extend MPC-Net, which is an Imitation Learning approach guided by Model Predictive Control (MPC).
The strategy of MPC-Net differs from many other approaches since its objective is to minimize the control Hamiltonian, which derives from the principle of optimality.
To represent the policies, we employ a mixture-of-experts network (MEN) and observe that the performance of a policy improves if each expert of a MEN specializes in controlling exactly one mode of a hybrid system, such as a walking robot.
We introduce new loss functions for single- and multi-gait policies to achieve this kind of expert selection behavior.
Moreover, we benchmark our algorithm against  Behavioral Cloning and the original MPC implementation on various rough terrain scenarios.
We validate our approach on hardware and show that a single learned policy can replace its teacher to control multiple gaits.
\end{abstract}
%
% Two or three meaningful keywords should be added here (only RAL)
%
%\begin{IEEEkeywords}
%Learning from Demonstration, Legged Robots, Optimization and Optimal Control
%\end{IEEEkeywords}
%\IEEEpeerreviewmaketitle % RAL
%

%
%===============================================================================
%

\section{Introduction}\label{sec:introduction}

The control of hybrid, underactuated walking robots is a challenging task, which becomes especially difficult in missions that require onboard real-time control.
In this scenario, training a feedback policy offline with demonstrations from \ac{OC} or \ac{MPC}~\cite{Mordatch2014, Kahn2017, Carius2020} is a promising option, as it combines the advantages of both data-driven and model-based approaches: a learned policy computes control inputs quickly online, while \ac{OC} provides a framework to find control inputs that respect constraints and optimize a performance criterion.

When a quadrupedal robot is deployed in different environments, it can be advantageous to adapt the gait.
For example, one might prefer a statically stable over a dynamically stable gait on uneven or slippery ground. 
However, in \ac{RL}, policies often converge to a single gait~\cite{Haarnoja2019, Hwangbo2019} with a few exceptions~\cite{Tan2018, Iscen2018}.
Further, works on \ac{IL} usually try to imitate one behavior per policy~\cite{Abbeel2010, Peng2020}, making transitioning between policies difficult, as for a walking robot, switching between policies for different behaviors can cause jerky and unstable locomotion~\cite{Iscen2018} unless one undesirably uses the stance mode for transitioning.
One solution is to use policy distillation to merge multiple task-specific policies into a single policy~\cite{Xie2019}.
An alternative is to add task-specific signals to the observation space of a generic policy~\cite{Iscen2018}.
Motivated by this alternative, our approach can train a single feedback policy that imitates multiple gaits from \ac{MPC} demonstrations and can switch between different gaits during execution.

Typically, \ac{IL} approaches can be categorized into two groups~\cite{Osa2018}: \ac{BC}~\cite{Bain1999} replicates the demonstrator's policy from state-input pairs without interacting with the environment, while \ac{IRL}~\cite{Abbeel2004} seeks to learn the demonstrator's cost or reward function, which is subsequently used to train a policy with a standard \ac{RL} procedure.
In contrast, our work is based on MPC-Net~\cite{Carius2020}, which is an \ac{IL} approach that uses solutions from \ac{MPC} to guide the policy search and attempts to minimize the control Hamiltonian, which also encodes the constraints of the underlying \ac{OC} problem.
MPC-Net differs from \ac{BC}, as our learner is never presented with the optimal control input, and from \ac{IRL}, as our cost function is obtained from a model-based controller instead of learning it.

The multi-modal nature of hybrid systems, such as walking robots, motivates the use of a \ac{MEN} architecture~\cite{Jacobs1991} for representing the control policy~\cite{Zhang2018, Carius2020}.
In this work, we provide further evidence that the performance of a policy improves if each expert of a \ac{MEN} specializes in controlling exactly one mode of a hybrid system.
This poses the challenge to reliably achieve such a single responsibility expert selection behavior, i.e., the \ac{MEN} has to partition the problem space such that a single expert is responsible for a distinct mode.
To this end, we introduce a new Hamiltonian loss function, which, compared to our previous work, leads to a better localization of the experts and, in turn, better performance on the robot.
Moreover, in more complex multi-gait settings, we show how a guided loss function provides a framework for directing the learner with domain knowledge towards advantageous expert selections.

\begin{figure}[t]
	\mbox{\includegraphics[width=0.325\columnwidth]{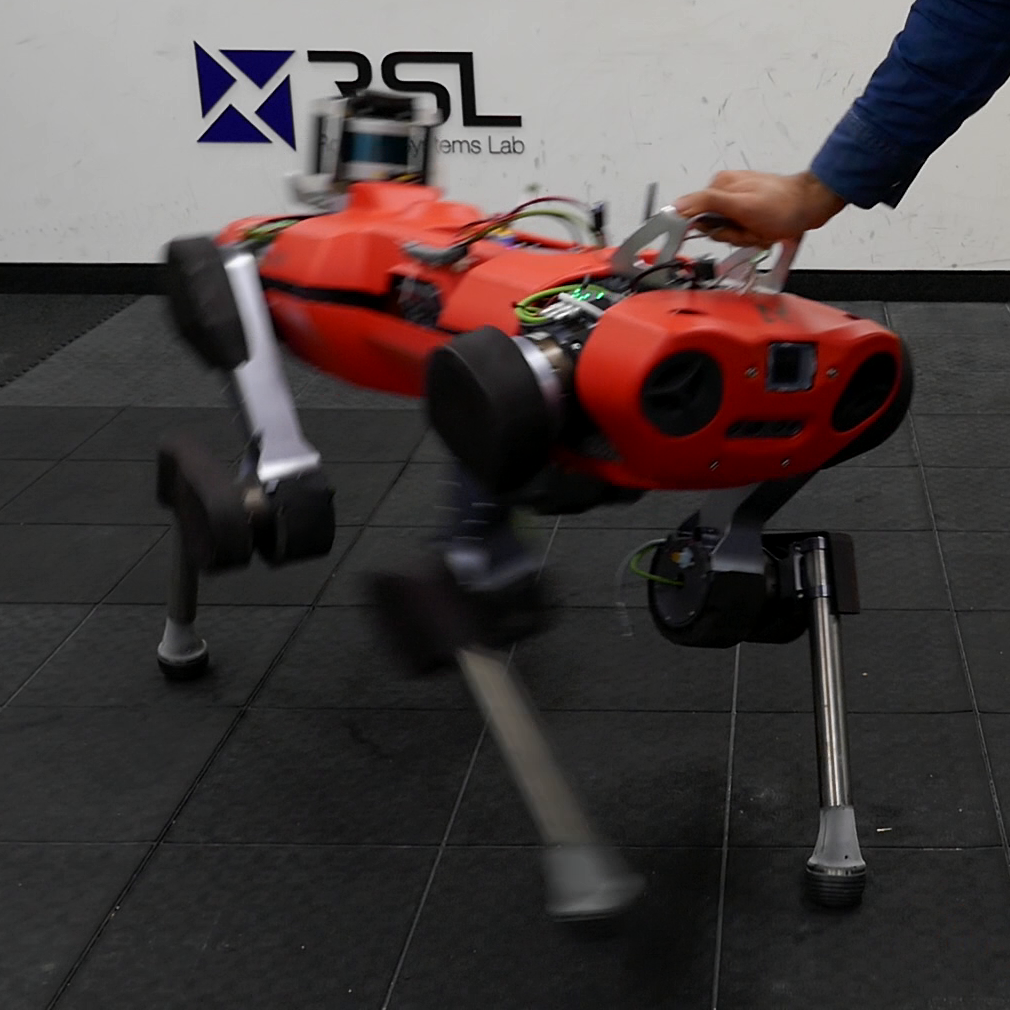}}
	\hfill
	\mbox{\includegraphics[width=0.325\columnwidth]{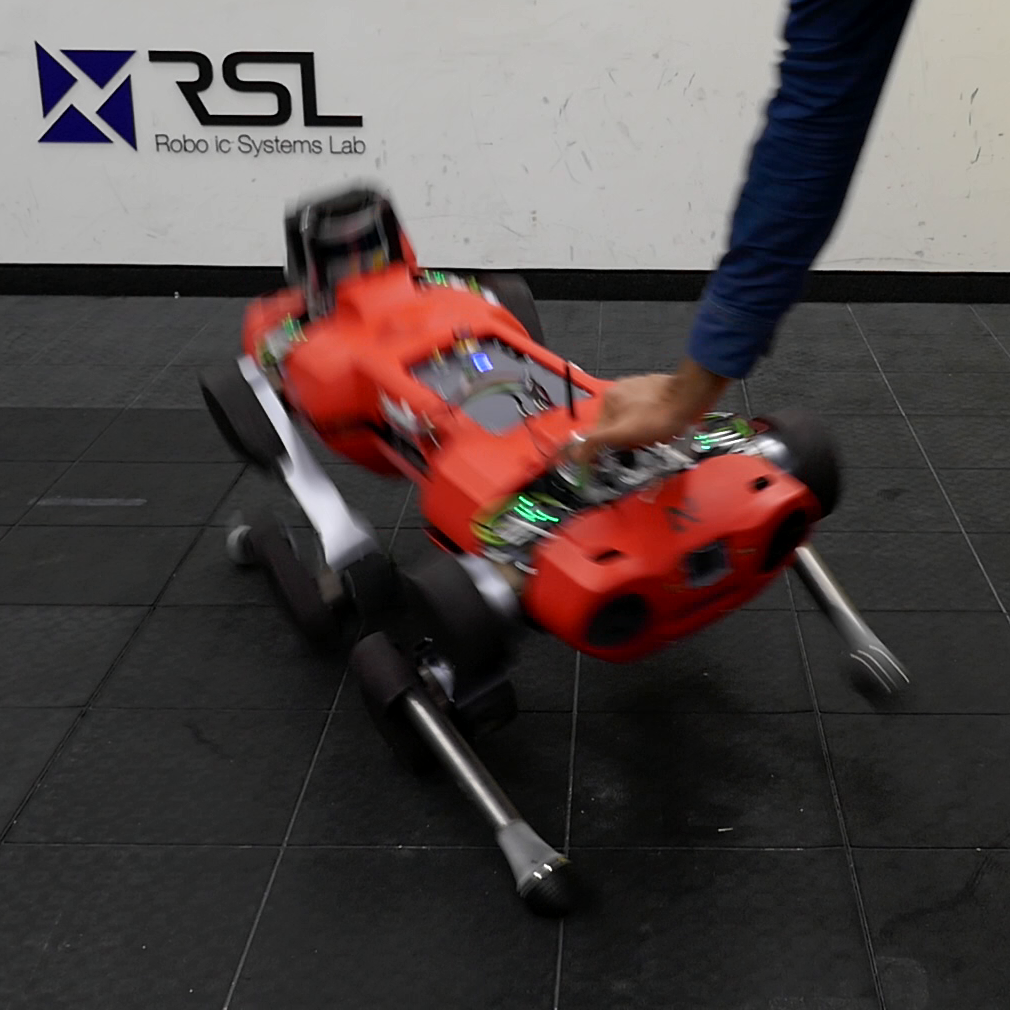}}
	\hfill
	\mbox{\includegraphics[width=0.325\columnwidth]{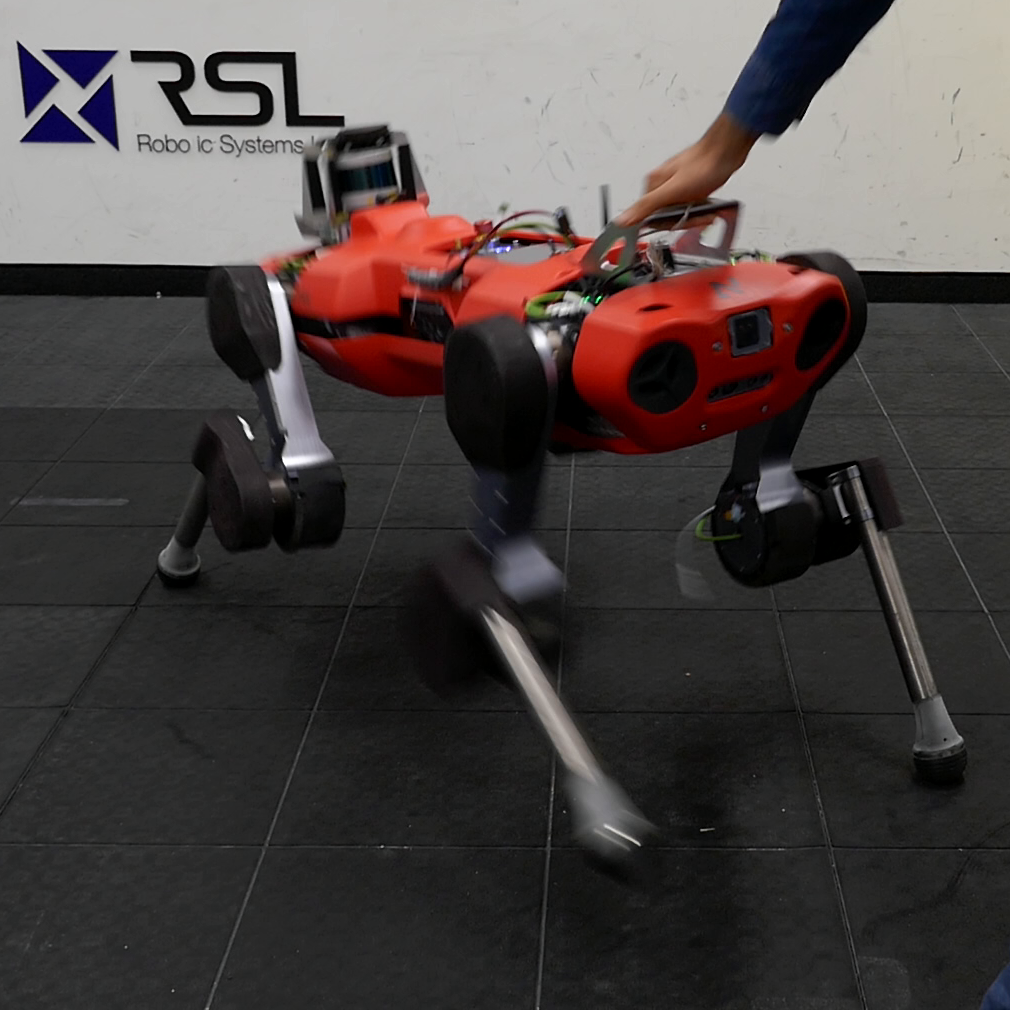}}
	\caption{ANYmal being pushed around while running a learned policy.}
	\label{fig:anymal}
\end{figure}

While our previous work~\cite{Carius2020} has established the underlying theoretical principle of a Hamiltonian loss function for policy search, this work contributes the following advances:
\begin{itemize}
\item Introduction of a new Hamiltonian loss function leading to a better localization of the \ac{MEN}'s experts (\ref{sec:L3}).
\item Proposal of a guided loss function that allows the incorporation of domain knowledge for an improved expert selection strategy (\ref{sec:cheating}).
\item Demonstration of the new Hamiltonian loss function resulting in improved performance (\ref{sec:ablation_study}).
\item Benchmarking experiments confirming that MPC-Net leads to more robust policies compared to \ac{BC} (\ref{sec:benchmarking}).
\item Results showing that a single policy can learn multiple gaits and execute them with high performance (\ref{sec:multi_gait}).
\end{itemize}

%
%===============================================================================
%

\section{Background}
This section covers the underlying control problem and recaps the main methodological aspects of the MPC-Net approach.

\subsection{Model Predictive Control}

\ac{MPC} provides solutions to the following \ac{OC} problem
\stepcounter{equation} % step equation counter and manually set correct tag for cost
\begin{subequations}
\begin{align}
& \underset{\vec{u}(\cdot)}{\text{minimize}}
& & \phi(\vec{x}(t_f)) + \int_{t_s}^{t_f} l(\vec{x}(t),\vec{u}(t),t) \de t, \label{eq:cost} \tag{1} \\
& \text{subject to} & & \vec{x}(t_s) = \vec{x}_s, \\
& & & \dot{\vec{x}} = \vec{f}(\vec{x},\vec{u},t), \\
& & & \vec{g}(\vec{x},\vec{u},t) = \vec{0}, \\
& & & \vec{h}(\vec{x},\vec{u},t) \geq \vec{0},
\end{align}
\label{eq:constraints}%
\end{subequations}
where $\vec{x}(t)$ and $\vec{u}(t)$ are the state and input at time $t$.
The objective is to minimize the final cost $\phi$ and the time integral of the intermediate cost $l$ over the receding time horizon~${t_h = t_f - t_s}$, where $t_s$ is the start time and $t_f$ is the final time.
The initial state $\vec{x}_s$ is given, and the system dynamics are determined by the system flow map~$\vec{f}$.
Moreover, the minimization is subject to the equality constraints $\vec{g}$ and the inequality constraints $\vec{h}$.

For the optimization, we use a \ac{SLQ} algorithm~\cite{Farshidian2017a}, which is a variant of the Differential Dynamic Programming (DDP) algorithm~\cite[p. 570]{Rawlings2017}.
The equality constraints $\vec{g}$ are handled using Lagrange multipliers~$\vec{\nu}$~\cite{Farshidian2017a}, and the inequality constraints $\vec{h}$ are considered through a barrier function $b$~\cite{Grandia2019b}.
Therefore, the corresponding Lagrangian is given by
\begin{equation}
\begin{split}
\mathcal{L}(\vec{x},\vec{u},t) = l(\vec{x},\vec{u},t) & + \vec{\nu}(\vec{x},t)^\top \vec{g}(\vec{x},\vec{u},t) \\ & + \sum_{i} b (h_i(\vec{x},\vec{u},t)).
\end{split}
\end{equation}

The solution to the OC problem~(\ref{eq:cost},~\ref{eq:constraints}) can be represented by a linear control policy based on the nominal state trajectory $\vec{x}_{\text{nom}}(\cdot)$, the nominal input trajectory $\vec{u}_{\text{nom}}(\cdot)$, and the time-varying linear feedback gains $\mat{K}(\cdot)$ as
\begin{equation}
\vec{\pi}_{\text{mpc}}(\vec{x},t) = \vec{u}_{\text{nom}}(t) + \mat{K}(t)(\vec{x} - \vec{x}_{\text{nom}}(t)).
\end{equation}
Moreover, the optimal cost-to-go function $V$ and the control Hamiltonian $\mathcal{H}$ can be written respectively as
\begin{align} 
&V(\vec{x},t) = \min_{\substack{\vec{u}(\cdot) \\ \text{s.t.} \, \eqref{eq:constraints}}} \left\lbrace \phi(\vec{x}(t_f)) + \int_{t}^{t_f} l(\vec{x}(\tau),\vec{u}(\tau),\tau) \de \tau \right\rbrace ,
\\
&\mathcal{H}(\vec{x},\vec{u},t) = \mathcal{L}(\vec{x},\vec{u},t) + \partial_{\vec{x}}V(\vec{x},t) \vec{f}(\vec{x},\vec{u},t),
\label{eq:hamiltonian}
\end{align}
where the Hamiltonian satisfies the well-known Hamilton-Jacobi-Bellman (HJB) equation for all $t$ and $\vec{x}$
\begin{equation} 
0 = \min_{\vec{u}} \left\lbrace \mathcal{H}(\vec{x},\vec{u},t) +\partial_{t}V(\vec{x},t) \right\rbrace.
\label{eq:hjb}
\end{equation}

\subsection{MPC-Net}\label{sec:mpc_net}

The main idea of MPC-Net~\cite{Carius2020} is to imitate \ac{MPC} by minimizing the Hamiltonian while representing the corresponding control inputs by a parametrized policy $\vec{\pi}(\vec{x},t;\vec{\theta})$.
From the \ac{SLQ} solver we have a local solution $V(\vec{x},t)$ for~\eqref{eq:hjb} as well as access to its state derivative $\partial_{\vec{x}} V(\vec{x},t)$ and the optimal Lagrange multipliers $\vec{\nu}(\vec{x},t)$.
Therefore, finding a locally optimal control policy simplifies to minimizing the right-hand-side of~\eqref{eq:hjb}~\cite[p. 430]{Bertsekas2017}.
This motivates the strategy for finding the optimal parameters $\vec{\theta}^*$, which is given by
\begin{equation}
\vec{\theta}^* = \arg \min_{\vec{\theta}} \mathbb{E}_{ \left\{ t, \vec{x} \right\} \sim \mathcal{P} } \left[ \mathcal{H} (\vec{x}, \vec{\pi} (\vec{x}, t ; \vec{\theta}), t)\right], \label{eq:startegy}
\end{equation}
where the distribution $\mathcal{P}$ encodes which areas of the time-state space are visited by an optimal controller.
The key training steps of this \ac{IL} approach are schematically shown in Fig.~\ref{fig:schematic} and are discussed in more detail in Sec.~\ref{sec:training}.

As the \ac{SLQ} solver provides a local approximation of the optimal cost-to-go function and of the optimal Lagrange multipliers, MPC-Net takes samples around the nominal state to augment the data set.
This reduces the number of \ac{MPC} calls needed to successfully train a control policy and makes the learned policy more robust.

To address the mismatch between the distributions of states visited by the optimal and learned policy~\cite{Ross2011}, MPC-Net forward-simulates the system with the behavioral policy
\begin{equation}
\vec{\pi}_{b} (\vec{x}, t ; \vec{\theta}) = \alpha \vec{\pi}_{\text{mpc}} (\vec{x}, t) + (1 - \alpha) \vec{\pi} (\vec{x}, t ; \vec{\theta}), 
\end{equation}
where the mixing parameter $\alpha$ linearly decreases from one to zero in the course of the training.

As mentioned in Sec.~\ref{sec:introduction}, for hybrid systems, it is beneficial to represent the policy by a \ac{MEN} architecture.
Therefore, the output of the network is the policy
\begin{equation}
\vec{\pi} (\vec{x}, t ; \vec{\theta}) = \sum_{i = 1}^{E} p_i (\vec{x}, t ; \vec{\theta}) \vec{\pi}_i (\vec{x}, t ; \vec{\theta}), \label{eq:policy}
\end{equation}
where the expert weights $\vec{p} = (p_1 , \dots , p_E )$ are calculated by a gating network and each expert policy $\vec{\pi}_i$ is computed by one of the $E$ experts.

Inserting~\eqref{eq:policy} into~\eqref{eq:startegy} leads to the loss function
\begin{equation}
L_1 = \mathcal{H} \left( \vec{x},\sum_{i = 1}^{E} p_i (\vec{x}, t ; \vec{\theta}) \vec{\pi}_i (\vec{x}, t ; \vec{\theta}), t \right). \label{eq:L1}
\end{equation}
However, this loss results in cooperation rather than competition between the experts.
To encourage expert specialization, it is better to force each expert to individually minimize the objective~\cite{Jacobs1991}.
For MPC-Net, this leads to the loss function 
\begin{equation}
L_2 = \sum_{i = 1}^{E} p_i (\vec{x}, t ; \vec{\theta}) \mathcal{H} (\vec{x}, \vec{\pi}_i (\vec{x}, t ; \vec{\theta}), t). 
\label{eq:L2}
\end{equation}

Training the policy requires the gradient of the loss function w.r.t. the parameters.
For better readability, we define
\begin{align}
    p_i &= p_i(\vec{x},t;\vec{\theta}), \\
    \mathcal{H}_i &= \mathcal{H}(\vec{x},\vec{\pi}_i(\vec{x},t;\vec{\theta}),t).
\end{align}
Then, the gradient of~\eqref{eq:L2} is
\begin{equation}
\frac{\partial L_2}{\partial \vec{\theta}}  = \sum_{i = 1}^{E} p_i \frac{\partial \mathcal{H}_i}{\partial \vec{u}} \frac{\partial \vec{\pi}_i}{\partial \vec{\theta}} + \mathcal{H}_i \frac{\partial p_i}{\partial \vec{\theta}}, \label{eq:gradient_L2}
\end{equation}
where the input derivative of the Hamiltonian $\partial_{\vec{u}} \mathcal{H}_i$ can be queried from the \ac{SLQ} solver, and the gradients $\partial_{\vec{\theta}} \vec{\pi}_i$ and~$\partial_{\vec{\theta}} p_i$ are computed by backpropagation.

%
%===============================================================================
%

\section{Method}

In this section, we first introduce the new Hamiltonian loss function and then the guided loss function.

\subsection{Log-Partitioned Loss Function}\label{sec:L3}

In a least-squares setting, Jacobs et al.~\cite{Jacobs1991} discuss a third \ac{MEN} loss function, which is the negative log probability of a Gaussian mixture model and reportedly results in a better performance.
Inspired by that, we introduce the loss function
\begin{equation}
L_3 = - \frac{1}{\beta} \log \left( \sum_{i = 1}^{E} p_i \exp \left( - \beta \left( \mathcal{H}_i + \partial_t V \right) \right) \right), \label{eq:L3}
\end{equation}
where $\partial_t V = \partial_{t}V(\vec{x},t)$ is a bias term motivated by~\eqref{eq:hjb} and computed by numerical differentiation, and $\beta$ is an inverse temperature parameter.
Note that the sum in the argument of the logarithm has some similarity to a partition function.

The expert weight $p_i$ can be seen as the prior probability that expert $i$ can minimize the Hamiltonian at the current observation.
In that light, we define the posterior probability
\begin{equation}
q_i = q_i(\vec{x},t; \vec{\theta}) := \frac{p_i \exp \left( - \beta \left( \mathcal{H}_i + \partial_t V \right) \right)}{\sum\limits_{j=1}^{E} p_j \exp \left( - \beta \left( \mathcal{H}_j + \partial_t V \right) \right)},
\end{equation}
which is a better estimation of the probability that expert $i$ can minimize the Hamiltonian.
Then, the gradient of~\eqref{eq:L3} can be written as
\begin{equation}
\frac{\partial L_3}{\partial \vec{\theta}} = \sum_{i = 1}^{E} q_i \frac{\partial \mathcal{H}_i}{\partial \vec{u}} \frac{\partial \vec{\pi}_i}{\partial \vec{\theta}} - \frac{1}{\beta} \frac{q_i}{p_i} \frac{\partial p_i}{\partial \vec{\theta}}. \label{eq:gradient_L3}
\end{equation}
Compared to~\eqref{eq:gradient_L2}, the expert updates are weighted by the posterior instead of the prior.
Moreover, notice the difference in the sign of the gating updates.
In~\eqref{eq:gradient_L2} the expert weight~$p_i$ receives a penalty given by the size of the corresponding Hamiltonian, whereas in~\eqref{eq:gradient_L3} the expert weight~$p_i$ receives a reward according to the ratio of the posterior and prior probabilities.
In Sec.~\ref{sec:ablation_study} we show that these properties indeed lead to improved performance.

To better understand the new loss function, note that we can also get~\eqref{eq:gradient_L3} from the gradient of
\begin{equation}
    \sum_{i = 1}^{E} \bar{q}_i \mathcal{H}_i + \frac{1}{\beta} \left( - \sum_{i = 1}^{E} \bar{q}_i \log (p_i) \right), \label{eq:interpretation_L3}
\end{equation}
where $\bar{q}_i$ is equal to $q_i$ but detached from the computational graph, and thus no gradient will be backpropagated along this variable.
The first term is similar to $L_2$ but the expert weight $p_i$ is replaced by $\bar{q}_i$.
The second term in parentheses is the cross-entropy $CE(\bar{q}, p)$ and pulls the prior towards the current posterior, which is fixed for the moment.

\subsection{Guided Loss Function}\label{sec:cheating}

To direct the learner with domain knowledge towards advantageous expert selections and influenced by the learning by cheating idea~\cite{Chen2019}, we propose the guided loss
\begin{equation}
L_G = L_E + \lambda L_D,
\end{equation}
where $L_E \in \{ L_1, L_2, L_3\}$ is the expert loss, $L_D$ is a loss that incorporates the domain knowledge, and the parameter~$\lambda$ controls the relative importance of both loss types.

In the context of hybrid systems, the expert selection should be related to the mode selection in \ac{OC}. 
The gait or mode selection either takes place based on optimization~\cite{Winkler2018} or based on other domain knowledge, such as qualification of the terrain~\cite{Brandao2020} or commanded speeds~\cite{Xi2016}.
In the simplest case, this leads to a mode schedule $m(t)$ that returns the active mode $i$ at any time $t$.
In general, however, the mode selection $m(\vec{x},t)$ can also depend on the current observed state~$\vec{x}$.
In that case, we define the empirical probability to observe mode $i$ as
\begin{equation}
\tilde{p}_i = \tilde{p}_i(\vec{x},t).
\end{equation}

To incorporate our domain knowledge that the experts and modes should match, we maximize the log-likelihood, which is equivalent to minimizing the cross-entropy~\cite[p. 129]{Goodfellow2016}.
Thus, for $L_1$ and $L_2$, $L_D$ is given by the cross-entropy 
\begin{equation}
CE(\tilde{p}, p) = - \sum_{i = 1}^{E} \tilde{p}_i \log (p_i). \label{eq:cheating_L1_L2}
\end{equation}
For $L_3$, eq.~\eqref{eq:cheating_L1_L2} and the cross-entropy term in~\eqref{eq:interpretation_L3} can be in conflict if the posterior does not agree with the observed modes.
To avoid this scenario, it is better to guide $L_3$ with
\begin{equation}
CE(\tilde{p}, q) = - \sum_{i = 1}^{E} \tilde{p}_i \log (q_i), \label{eq:cheating_L3}
\end{equation}
which, with the interpretation in~\eqref{eq:interpretation_L3}, encourages the predicted expert selection to match the observed mode selection.

In the \ac{MEN} literature, one can distinguish between a \ac{MILE} and a \ac{MELE}~\cite{Masoudnia2014}.
\ac{MILE} uses a competitive process to localize the experts.
For example, $L_2$ and $L_3$ belong to this group.
In contrast, \ac{MELE} assigns the experts to pre-specified clusters.
In a broader sense, the guided loss $L_G$ can be seen as part of the \ac{MELE} group.
However, note that our method is different from hardcoding the assignment of the experts according to some heuristic and training separate networks for each case, as we allow the modes and their probabilities to be a function of the observations.
Therefore, in this framework, the gating network can learn to deviate from a heuristic, such as a mode schedule for a hybrid system, and adapt the expert selection to the current observations.
The benefits of the guided loss idea become evident in Sec.~\ref{sec:multi_gait}.

Finally, it should be noted that a limitation of the presented methodology is that it requires a mode schedule. To address this limitation, one could try to learn a state-based mode selection $m(\vec{x})$, similar to works that predict mode schedules from states to facilitate solving the \ac{OC} problem~\cite{Hogan2020}.

%
%===============================================================================
%

\section{Implementation}

% belongs to \subsection{Policy Architecture} but inserted here for the layout
\begin{figure}[t]
    \centering
    \includegraphics[width=0.9\columnwidth]{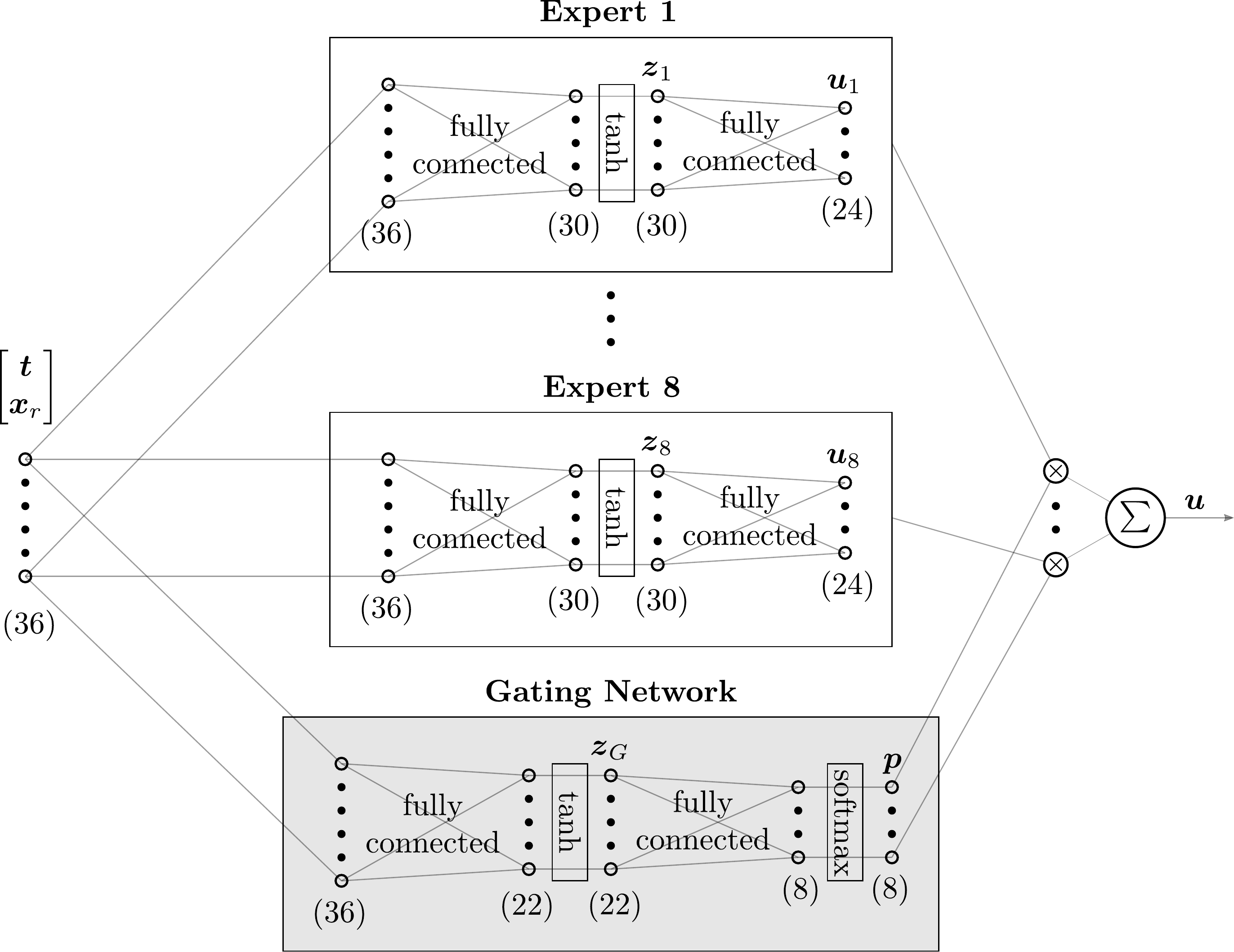}
    \caption{\ac{MEN} architecture instantiated for the ANYmal robot.}
    \label{fig:architecture}
\end{figure}

This section presents how the method is applied to a quadrupedal robot.
Moreover, we explain our policy architecture, the training procedure, and the deployment pipeline.

\subsection{ANYmal Control}

In this work, we use the quadrupedal robot ANYmal (Fig.~\ref{fig:anymal}), which is a hybrid system due to discrete switches in the contact configuration.
The feet are constrained to have zero contact forces in the swing phase and zero velocity in the contact phase.
For \ac{MPC}, the robot is represented by a kinodynamic model, which has $24$ states (base pose, base twist, joint positions) and $24$ inputs (foot contact forces, joint velocities) \cite{Farshidian2017a}.
The \ac{OC} cost function~\eqref{eq:cost} is determined by
\begin{align}
\phi(\vec{x}) &= (\vec{x} - \vec{x}_d(t_f))^\top \mat{Q}_f (\vec{x} - \vec{x}_d(t_f)), \\
l(\vec{x},\vec{u},t) &= (\vec{x}-\vec{x}_d(t))^\top \mat{Q} \, (\vec{x}-\vec{x}_d(t)) + \vec{u}^\top \mat{R} \, \vec{u},
\end{align}
where $\vec{x}_d(\cdot)$ is a desired state trajectory that should be tracked.
We consider two gaits: trot, moving the diagonal legs together, and static walk, moving one leg at a time.
So, including the stance mode, we have to control $M=7$ modes.

\subsection{Gait Parametrization}\label{sec:gait_parametrization}

Based on the absolute time and the current state, it is difficult for the learned policy to infer which legs should be moved.
Therefore, it is more direct to provide a parametrization of the gait.
From the mode schedule we extract the leg phases $\vec{\varphi} = ( \varphi_{LF}, \varphi_{RF}, \varphi_{LH}, \varphi_{RH})$ according to
\begin{equation}
\varphi_i =
\begin{cases}
\displaystyle\frac{t - t_{lo}}{t_{td} - t_{lo}}, & \text{if leg $i$ in swing},\\
0, & \text{if leg $i$ in contact},
\end{cases}
\end{equation}
where $t_{lo}$ is the liftoff and $t_{td}$ the touchdown time.
By abuse of notation, we replace the absolute time $t$ in the policy~\eqref{eq:policy} with the so-called generalized time
\begin{equation}
\vec{t} = \begin{bmatrix} \vec{\varphi} & \vec{\dot{\varphi}} & \sin(\pi \vec{\varphi}) \end{bmatrix}^\top.
\end{equation}
The sinusoidal bumps $\sin(\pi \vec{\varphi})$~\cite{Carius2020} provide a reference for the swing motion.
We add the leg phases $\vec{\varphi}$ to learn asymmetries between the liftoff and touchdown phase and their time derivatives $\vec{\dot{\varphi}}$ to capture different swing speeds.
We address the benefits of this parametrization in Sec.~\ref{sec:ablation_study}.

\subsection{Relative State}\label{sec:relative_state}

For reference tracking, we replace the state $\vec{x}$ in the policy~\eqref{eq:policy} with a tracking error called the relative state
\begin{equation}
\vec{x}_r(t) = \mat{T}(\vec{\theta}_B) \left( \vec{x} - \vec{x}_d(t) \right),
\end{equation}
where $\vec{\theta}_B$ is the current orientation of the base in the world frame and the matrix $\mat{T}(\vec{\theta}_B)$ transform the pose error from the world into the base frame to make the policy training and deployment invariant w.r.t. the absolute orientation.

\subsection{Policy Architecture}

The \ac{MEN} architecture for the policies is shown in Fig.~\ref{fig:architecture}.
We use $E=8$ \ac{MLP} experts and a \ac{MLP} gating network with a softmax output activation, which ensures that the expert weights $\vec{p}$ are positive and sum to one.
Note that as long as $E \geq M$, training and deployment are not sensitive w.r.t. the parameter $E$, and the gating network is able to learn to select an appropriate expert for the mode~\cite{Carius2020}.
For example, we also trained policies with~$E=12$ experts but could not observe an advantage in terms of expert selection or policy performance.

Carius et al.~\cite{Carius2020} use an architecture with a common hidden layer for linear experts and a linear gating network, which has a sigmoid output activation with a subsequent normalization.
While we postulate that the common hidden layer is beneficial for training from little demonstration data, we use several \ac{MLP} experts since this allows the expert networks to extract distinct features for the individual modes that they are responsible for.
In combination with the loss function $L_2$, the normalized sigmoids help to select a consistent number of experts~\cite{Carius2020}.
However, the more common softmax better corresponds to the concept of single responsibility, and we handle the issue of expert selection with the newly introduced loss functions $L_3$ and $L_G$.

\subsection{Training} \label{sec:training}

\begin{figure}[t]
    \centering
    \includegraphics[width=0.9\columnwidth]{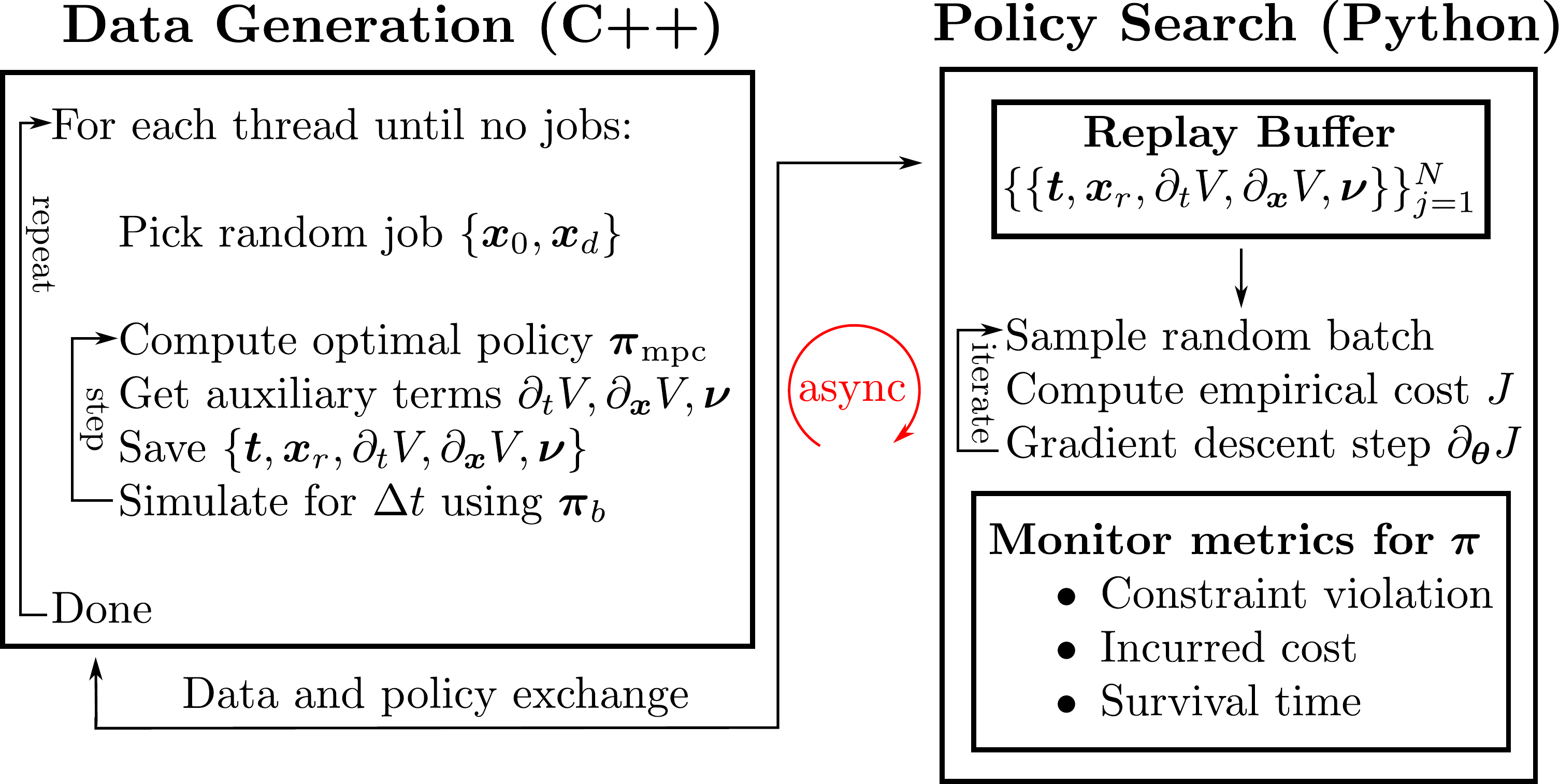}
    \caption{Schematic of the MPC-Net training procedure.}
    \label{fig:schematic}
\end{figure}

The training procedure is schematically shown in Fig.~\ref{fig:schematic}.
First, note that the multi-threaded data generation and the policy search run asynchronously.
While MPC-Net can stabilize different gaits from less than \unit[10]{min} of demonstration data~\cite{Carius2020}, the multi-threaded data generation ensures that the amount of data is not a bottleneck in this work.

Data are generated by $n_t$ threads that work on $n_j$ jobs per data generation run.
For each job, we start from a random initial state $\vec{x}_0$ with the task to reach a desired state $\vec{x}_d$ within the rollout length $T$.
In a loop, we run \ac{MPC}, save the data, and forward simulate the system by the time step $\Delta t$.
To avoid storing data that have similar informational content, we downsample the data and thus only save the nominal data and the data from $n_s$ samples around the nominal state in every $d_d$-th step.
A rollout is considered as failed, and its data are discarded if the pitch or roll angle exceeds \unit[30]{$^\circ$} or if the height deviates more than \unit[20]{cm} from the default value.

If a data generation run has completed, the data are pushed into the learner's replay buffer of size $N$.
In every learning iteration, we draw a batch of $B$ tuples $\{ \vec{t},\vec{x}_r, \partial_t V, \partial_{\vec{x}} V, \vec{\nu} \}$ from the replay buffer, compute the empirical cost
\begin{equation}
J(\vec{\theta}) = \frac{1}{B} \sum_{j = 1}^{B} L(\vec{t}_j, \vec{x}_{r,j}, \partial_t V_j, \partial_{\vec{x}} V_j, \vec{\nu}_j, \vec{\theta}),
\end{equation}
where $L \in \{ L_1, L_2, L_3, L_G \}$, and perform a gradient descent step in the parameter space using the Adam optimizer~\cite{Kingma2015}.

To monitor the training progress and as a substitute for a validation set, we perform a rollout with the learned policy at every $d_m$-th iteration and compute the following metrics: the average constraint violation, the incurred cost~\eqref{eq:cost}, and the survival time until our definition of failure applies.

When using the system dynamics $\vec{f}$ for the rollouts~\cite{Carius2020}, it is possible that the learned policy cheats, e.g., by applying forces in mid-air.
In this work, we use the physics engine RaiSim~\cite{Hwangbo2018} for the data generation and metrics rollouts.
The availability of more realistic training data improves the sim-to-real transfer, and using a physics engine for validation leads to more accurate metrics of true performance.
The mentioned hyperparameters are summarized in Table~\ref{tab:hyperparameters}.

\begin{table}[t]
    \centering
    \caption{Hyperparameters of MPC-Net.}
    \begin{tabularx}{1.0\columnwidth}{rl||rl}
        \toprule
        time step $\Delta t$ & \unit[0.0025]{s} & inverse temperature $\beta$ & 1.0 \\
        rollout length $T$ & \unit[4]{s} & guided loss weight $\lambda$ & 1.0 \\
        number of threads $n_t$ & 5 & number of experts $E$ & 8\\
        number of jobs $n_j$ & 10 & batch size $B$ & 32 \\
        number of samples $n_s$ & 1 & learning rate $\eta$ & 1e-3 \\
        data decimation $d_d $& 4 & iterations single-gait $i_s$ & 100k \\
        metrics decimation $d_m$ & 200 & iterations multi-gait $i_m$ & 200k \\
        \bottomrule
    \end{tabularx}
    \label{tab:hyperparameters}
\end{table}

\subsection{Deployment}

\begin{figure}[b]
    \centering
    \includegraphics[width=0.9\columnwidth]{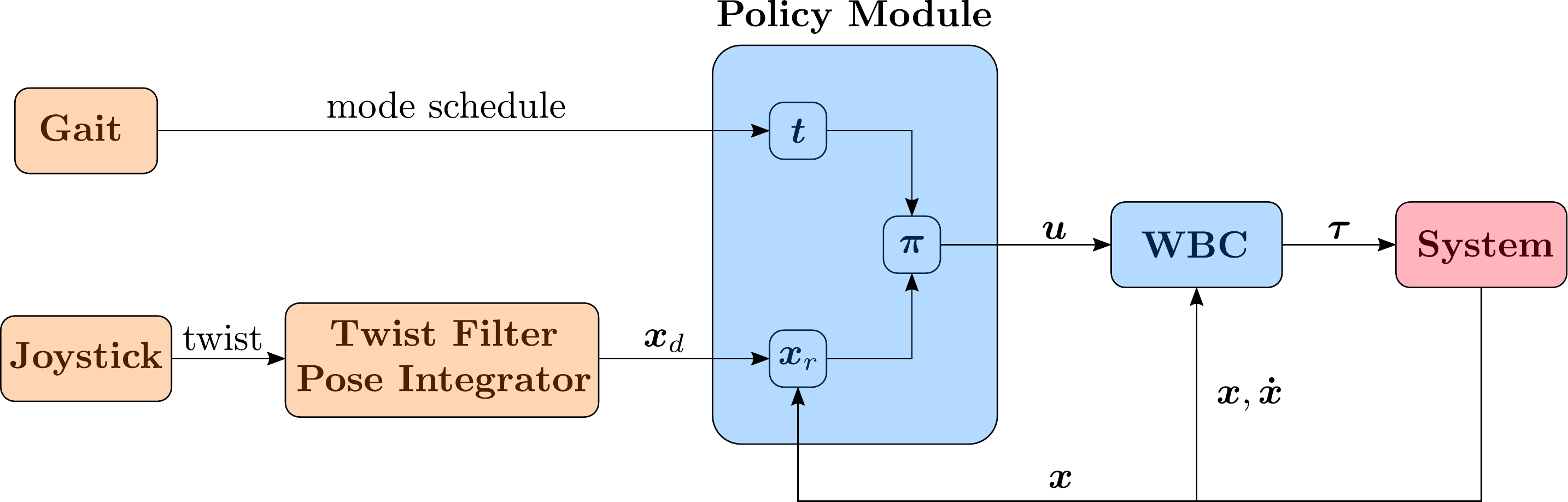}
    \caption{Deployment pipeline with the policy module replacing \ac{MPC}.}
    \label{fig:pipeline}
\end{figure}

Simply put, the learned policy replaces \ac{MPC} in our control architecture.
More specifically, our controller is given a mode schedule, the current state~$\vec{x}$, and a desired state~$\vec{x}_d$, which is generated from the user's commands (Fig.~\ref{fig:pipeline}).
From these quantities, the relative state~$\vec{x}_r$ and the generalized time~$\vec{t}$ can be assembled and passed into the policy network.
The kinodynamic control inputs $\vec{u}$, which are inferred from the learned policy $\vec{\pi}$ using ONNX Runtime~\cite{ONNXRuntime}, are tracked by a \ac{WBC} that computes the actuator torque commands $\vec{\tau}$ \cite{Bellicoso2016}.

%
%===============================================================================
%

\section{Results}

In this section, we show how our methodological contributions and implementation details lead to improved results compared to prior work.
To assess the performance of a learned policy, we found that the survival time and the constraint violation computed from the metrics rollouts are good indicators.
For the actual performance on hardware, we refer to the supplementary video\footnote{Link: \texttt{https://youtu.be/AUNIhr5I6Dg}}.
In the presented plots, noisy data, e.g., from the metrics rollouts, are filtered by an exponential moving average filter with smoothing factor $0.9$.

\begin{figure*}[t]
    \centering
    \subfloat[]{\includegraphics[width=0.335\textwidth]{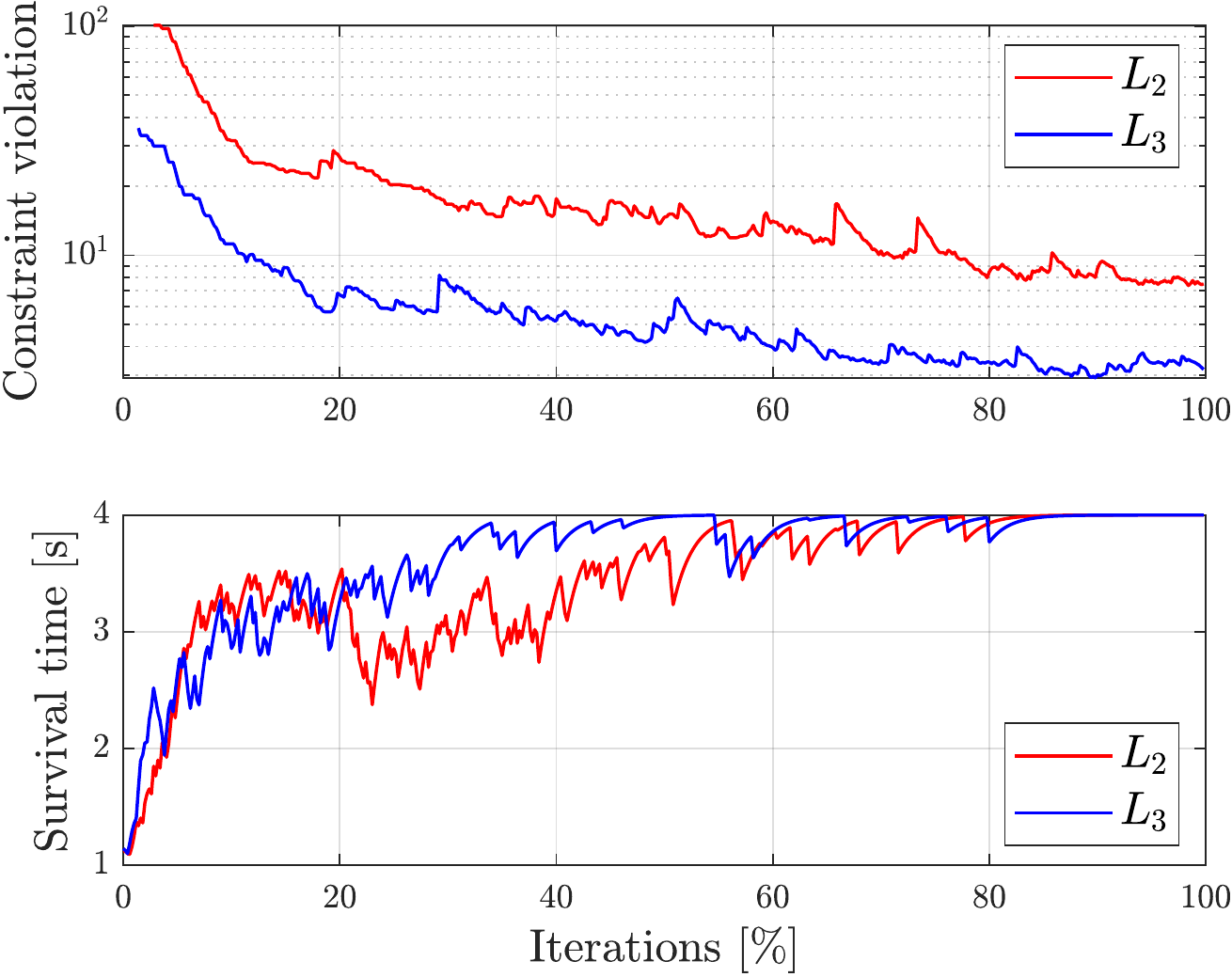}
    \label{subfig:ablation_study}}
    \hfil
    \subfloat[]{\includegraphics[width=0.315\textwidth]{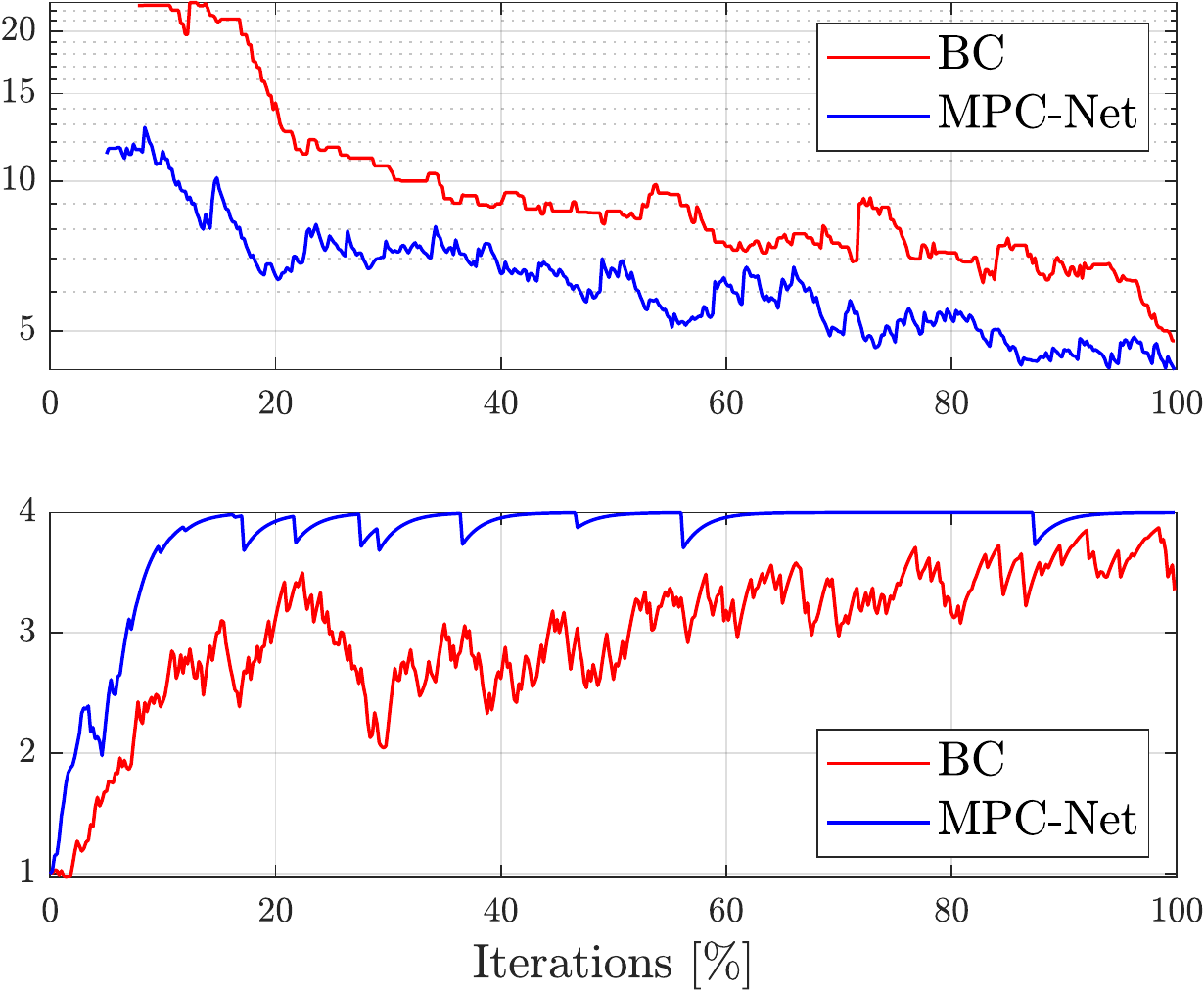}
    \label{subfig:benchmarking}}
    \hfil
    \subfloat[]{\includegraphics[width=0.32\textwidth]{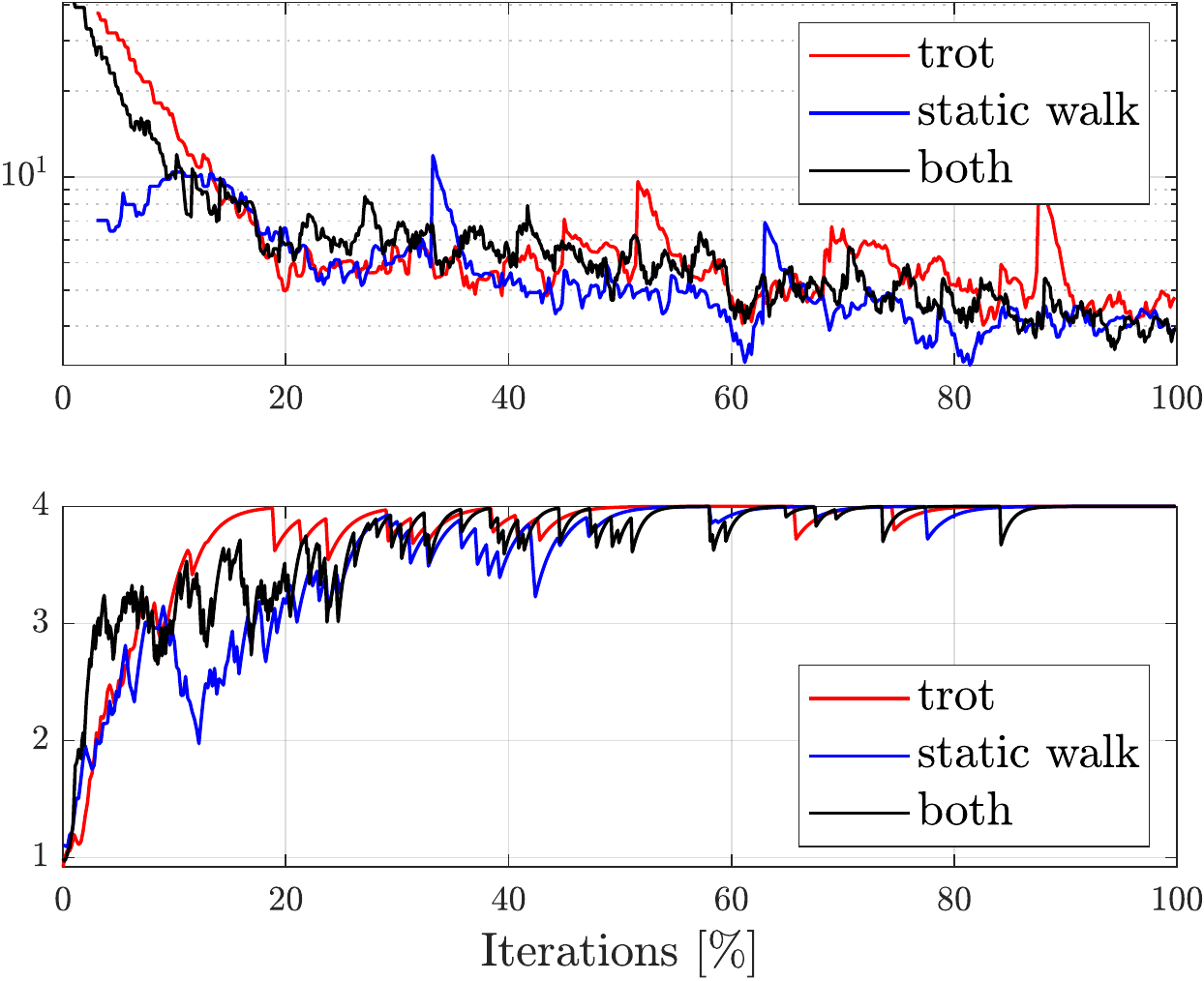}
    \label{subfig:multi_gait}}
    \vspace{5pt}
    \caption{The top graphs show a comparison of the constraint violation and the bottom graphs a comparison of the survival time. In this figure, all policies trained with loss $L_3$ achieve single responsibility of the experts for the modes. (a)~Performance of static walk when training with loss $L_2$ or~$L_3$. (b)~Performance of trot when using the \ac{BC} or MPC-Net training approach. (c)~Performance of a multi-gait policy consisting of trot and static walk compared to the corresponding single-gait policies. Note that the multi-gait training goes through twice as many iterations, and the policies can also control the pose.}
    \vspace{-13pt}
    \label{fig:subfig}
\end{figure*}

\subsection{Ablation Study}\label{sec:ablation_study}

We begin by providing evidence that the generalized time~$\vec{t}$ and the new loss $L_3$ help the MPC-Net algorithm to find better policies compared to our previous work~\cite{Carius2020}.

Extending the gait parameterization $\sin(\pi \vec{\varphi})$ with $\vec{\varphi}$ and~$\vec{\dot{\varphi}}$ reduces the constraint violation by a factor of four.
As the~$\vec{\dot{\varphi}}$ are rectangular functions, we conjecture that they facilitate changing experts at mode switches and, ergo, learning to respect the constraints of the swing and contact phases.

In Fig.~\ref{subfig:ablation_study} we compare the loss $L_2$ with the new loss~$L_3$, which achieves a lower constraint violation as well as a faster increase in the survival time.
Note that in Fig.~\ref{subfig:ablation_study} the policy trained with the loss $L_2$ only employs three experts, where one expert is responsible for the stance and two static walk modes.
Based on many experiments and as indicated by Fig.~\ref{subfig:ablation_study}, we conclude that the performance of a policy improves if each expert of a \ac{MEN} specializes in controlling exactly one mode of a hybrid system.
Table~\ref{tab:single_responsibility} shows that the new loss $L_3$ is better at achieving this single responsibility expert selection in general and especially for our choice of the inverse temperature $\beta = 1.0$.
We observe that the gating network commits to a certain expert selection early in the training process, which highlights the importance of efficient gradient updates as given by~\eqref{eq:gradient_L3}.

\begin{table}[t]
    \centering
    \caption{Ability of the loss $L_2$ and $L_3$ to achieve single responsibility of the experts for the modes of trot and static walk. The shown percentages are based on ten training runs each.}
    \begin{tabularx}{0.8\columnwidth}{r|c|ccc}
         & $L_2$ & \multicolumn{3}{c}{$L_3$} \\
         & & $\beta = 0.5$ & $\beta = 1.0$ & $\beta = 2.0$ \\
        \toprule
        trot & 40\% & \textbf{100\%} & \textbf{100\%} & 50\% \\
        static walk & 0\% & 40\% & \textbf{90\%} & 70\% \\
        \bottomrule
    \end{tabularx}
    \label{tab:single_responsibility}
\end{table}

\subsection{Benchmarking}\label{sec:benchmarking}

We benchmark our algorithm against the original \ac{MPC} and \ac{BC}. For the latter, we replace the Hamiltonian and the bias term in~$L_3$ with the normalized mean-squared error from the model predictive controller's control commands. 

In Fig.~\ref{subfig:benchmarking} we compare the MPC-Net algorithm with \ac{BC}, whose policies do not persistently reach the desired survival time of \unit[4]{s}.
Note that only completed metrics rollouts are considered in the constraint violation plots.
In this respect, policies trained with \ac{BC} attain a slightly larger constraint violation for the rollouts that they survive but fail for rollouts with apparently more difficult tasks~${\left\lbrace \vec{x}_0, \vec{x}_d \right\rbrace}$ that only policies trained with MPC-Net complete successfully.

To quantify the robustness of the approaches, we deploy MPC as well as policies trained with MPC-Net and BC on rough terrain in RaiSim, command them to walk forward for at most \unit[20]{s}, and measure the survival time.
Table~\ref{tab:rough_terrain} shows that policies trained with MPC-Net outperform those trained with \ac{BC} in surviving rough terrain, which was not part of the training data.
While the effects are difficult to isolate, we imagine MPC-Net shows comparatively greater robustness by learning from the Hamiltonian, which also encodes the constraints that ensure physical feasibility.
Compared to MPC, MPC-Net achieves similar survival times.
We think that learning from \ac{MPC} simulated at \unit[400]{Hz} enables the policy to compete with MPC running at \unit[40]{Hz}, which is an achievable onboard update frequency for MPC on ANYmal.

\begin{table}[t]
    \centering
    \caption{Survival time (mean and standard deviation) when deploying MPC, MPC-Net, and BC policies on terrain with different scales of roughness based on fifty test runs each.}
    \begin{tabularx}{0.8\columnwidth}{c|c|c|c}
        z-scale~\cite{Hwangbo2018} & MPC & MPC-Net & BC \\
        \toprule
        0.0 & 20.0 $\pm$ 0.0 & 20.0 $\pm$ 0.0 & 20.0 $\pm$ 0.0 \\
        4.0 & 19.2 $\pm$ 2.6 & 18.9 $\pm$ 2.9 & 6.8 $\pm$ 4.5 \\
        8.0 & 12.2 $\pm$ 5.6 & 10.6 $\pm$ 6.3 & 2.6 $\pm$ 1.6 \\
        \bottomrule
    \end{tabularx}
    \label{tab:rough_terrain}
\end{table}

\subsection{Multi-Gait Policy}\label{sec:multi_gait}

With our approach, a single policy can learn multiple gaits and execute them with high performance.
Fig.~\ref{subfig:multi_gait} shows that a multi-gait policy achieves a constraint violation and a survival time that are similar to the corresponding single-gait policies.
Also, on hardware, we observe similar performance.

\begin{figure}[t]
    \centering
    \includegraphics[width=0.9\columnwidth]{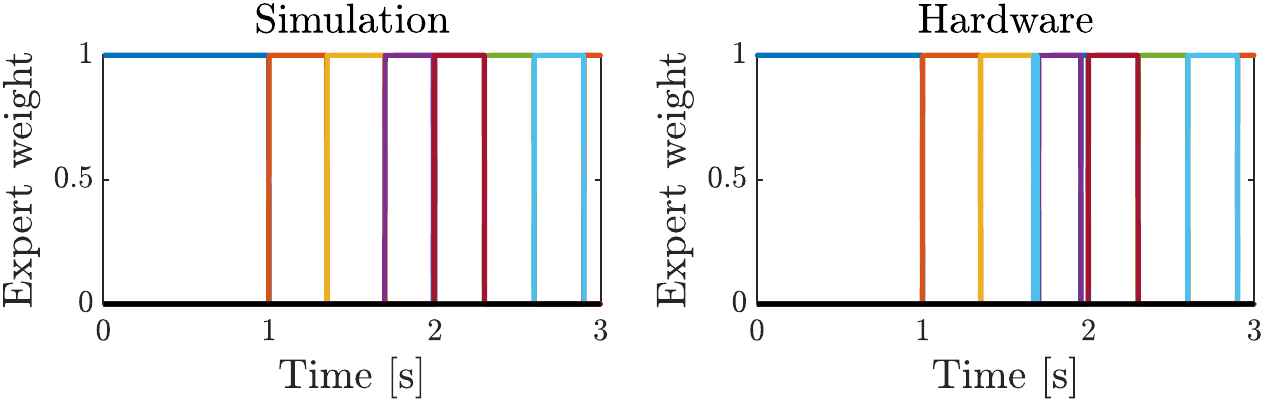}
    \caption{Expert selection in simulation (left) and on hardware (right) for a multi-gait policy trained with the guided loss $L_G$. The robot starts in stance (blue) and then alternates between trot (orange, yellow) and static walk (purple, red, green, light blue). One expert (black) is not used.}
    \label{fig:transition}
\end{figure}

Unfortunately, in multi-gait scenarios, $L_3$ does not reliably achieve single responsibility.
We observe that the network tends to commit too early to too few experts and conjecture that the Hamiltonian does not provide enough discrimination to reliably identify all modes.
Therefore, we propose to encode the single responsibility with the guided loss $L_G$.
In that case, the guided versions of $L_1$, $L_2$, and $L_3$ all achieve single responsibility and thus similar performance to the multi-gait policy shown in Fig.~\ref{subfig:multi_gait}.
Fig.~\ref{fig:transition} shows the expert selection for a policy trained with the guided loss~$L_G$.
In simulation, one can see that the expert selection corresponds to the mode schedule due to the single responsibility.
On hardware, the expert selection slightly deviates from the plan.
For example, at the end of the second trot mode (yellow), one leg is in early contact, which activates one of the static walk experts (light blue) for a short moment.

While we deem trot and static walk to be the practically most relevant gaits, we tested in simulation how our method scales to more than two gaits.
As shown in the video, we added a more exotic gait, namely dynamic diagonal walk, i.e., a hybrid of static walk and trot, to the multi-gait training.
In general, it can be noted that if the deployment deviates too much from the training data and especially if the execution requires new modes, such as for pace or bounding, then one has to explicitly train it and ensure that there are enough experts to cover all the modes of the gaits.

%
%===============================================================================
%

\section{Conclusion}

In this work, we observed that the performance of a policy improves if each expert of a \ac{MEN} specializes in controlling exactly one mode of a hybrid system.
Motivated by this, we introduced a new loss function, which leads to better expert localization and thus almost always achieves single responsibility for single-gait policies.
We showed that MPC-Net policies are comparatively robust.
Moreover, our method and implementation enable a single policy to learn multiple gaits by the incorporation of domain knowledge through a guided loss function.
Finally, we validated our approach on hardware and showed that the learned policies can replace \ac{MPC} during deployment.
This opens the door for more complicated scenarios that do not run in real time with \ac{MPC}. 

%
%===============================================================================
%

% 
%===============================================================================
% balance columns on final page
%\addtolength{\textheight}{-15cm}
%
\bibliographystyle{IEEEtran}
\bibliography{references} % .bib
\end{document}